\documentclass[runningheads]{llncs}
\usepackage{graphicx}
\usepackage{subcaption}
\usepackage{wrapfig}
\usepackage{xcolor}
\usepackage{array}
\usepackage{float}
\usepackage{theorem}
\usepackage{textfit}
\usepackage{placeins}
\usepackage{framed, color}
\usepackage{url}
\usepackage{rotating}
\usepackage{times}
\usepackage{amsmath}
\usepackage{amsfonts}
\usepackage{algpseudocode}
\usepackage{algorithm}
\usepackage{tabularx}
\usepackage{booktabs}
\usepackage{ragged2e}
\usepackage{listings}
\usepackage[breaklinks]{hyperref}
\usepackage{tabto}
\usepackage{lipsum}
\usepackage[export]{adjustbox}
\usepackage{caption}
\usepackage{multirow}
\usepackage{enumitem}
\usepackage{xcolor}
\usepackage[justification=centering]{caption}

\usepackage{listings}
\begin{document}
\title{An Explainable Disease Surveillance System for Early Prediction of Multiple Chronic Diseases}
\titlerunning{Chronic Diseases Early Prediction}
%
\author{Shaheer Ahmad Khan\inst{1}\orcidID{0009-0000-8772-8149}  \and Muhammad Usamah Shahid\inst{1}\orcidID{0009-0001-4293-2979}  
\and
Ahmad Abdullah\inst{2}
\and
Ibrahim Hashmat\inst{1}\and
Muddassar Farooq\inst{1}}
\authorrunning{S. A. Khan et al.}
%
\institute{CureMD Research, 80 Pine St 21st Floor, New York, NY 10005, United States
\email{\{shaheer.ahmed, muhammad.usamah, muddassar.farooq\}@curemd.com}
\and CureClinic, 30 Davis road, Lahore, Pakistan
\newline\url{https://www.curemd.com/}}

\maketitle              
\begin{abstract}
This study addresses a critical gap in the healthcare system by developing a clinically meaningful, practical, and explainable disease surveillance system for multiple chronic diseases, utilizing routine EHR data from multiple U.S. practices integrated with CureMD’s EMR/EHR system. Unlike traditional systems -- using AI Models that use features from patients labs -- our approach focuses on routinely available data -- like medical history, vitals, diagnoses, and medications -- to preemptively assess the risks of chronic diseases in the next year. We trained three distinct models for each chronic disease: prediction models that forecast the risk of a disease 3, 6, and 12 months before a potential diagnosis. We have developed Random Forest models, which were internally validated using the F1 scores and AUROC as the performance metrics, and were further evaluated by a panel of expert physicians for clinical relevance based on the inference grounded in the medical knowledge. Additionally, we discuss our implementation of integrating these models into a practical EMR system. Beyond using Shapely attributes and surrogate models for explainability, we also introduce a new rule engineering framework to enhance the intrinsic explainability of Random Forests.

\end{abstract}
\section{Introduction}
\subsection{Background} Chronic Diseases as defined by the NCI are conditions that last for more than 3 months and can typically be only controlled but not cured. According to the CDC, 60\% of adults in the US have at least one chronic disease and 40\% have two or more. Chronic Diseases are the leading cause of mortality and disability in the US. 90\% of \$4.1 trillion of US healthcare expenditures are spent on people with Chronic Diseases and Mental health conditions.

In this paper, we use Electronic Health Records (EHR), which hold a large amount of patients' healthcare data including demographics, vitals, diagnoses, and medications. This data is routinely collected during periodic clinical encounters. Consequently, AI models that only use these types of information elements to do an early prediction of the risk of chronic diseases, have a significant intrinsic value and utility. 

While research studies have been done to predict the risk of multiple chronic diseases as a group \cite{yang2021multiplechronic} \cite{may2020intermountain}, but predicting the risk of each individual disease of multiple chronic diseases, using a single AI system for real world clinical settings is a serious gap in the healthcare \cite{yang2021multiplechronic} \cite{bardhan2020multiplechronic}. The authors of \cite{hajat2018mccburden} highlight that 16-57\% adults in developed countries suffer from more than one chronic condition (40\% in the US). Moreover, each additional chronic condition exponentially increases the health care cost. Therefore, it is imperative to shift from a single disease risk surveillance framework to a surveillance system that is capable of predicting the risk of multiple chronic conditions. One study presented in \cite{junaid2022mccann} deals with multiple individual diseases, but this study has some limitations. Firstly, it does not report F Scores which are the recommended performance metric for imbalanced datasets. Secondly, the sample size is very small which hampers the generalizability of the inference engine. Thirdly, the authors admit that this approach does not scale to real-world clinical applications, as their model uses curated feature sets that are not available in real-world EHR systems. Last but not the least, they also use lab tests (which limits their applicability in real-world settings) as lab investigations are only done to confirm a diagnosis when physicians have already suspected the presence of a chronic disease.

It is worth mentioning that lab results are used for chronic disease diagnosis for example in \cite{hussein2012cddlabs}, the authors use HbA1c to predict Diabetes, or the use of ``TSH" levels as a feature in \cite{chaganti2022thyroidlabs} to predict hypothyroidism, or the use of LDL and HDL levels to predict dyslipidemia \cite{gutierrez2023dylipidemialabs}. Multiple lab test values like sodium, potassium, serum creatinine etc. have been used to predict Chronic Kidney Disease \cite{chittora2021kidneylabs}. For predicting early heart disease, blood sugar levels and serum cholesterol are routinely used \cite{mohan2019heartlabs}. Similarly, for predicting Hypertension, labs are also used \cite{ye2018hypertensionlabs}. The lab test results are typically not available at the time of clinical decision-making \cite{pincus2009labs} for the majority of patients, which substantially reduces sample size for training. Doing multiple imputations on the data with a large proportion of missing values (about 67\%) can introduce a bias \cite{boursi2016labs}. Therefore, we excluded the lab results from our study.

XAI (Explainable AI) in medicine is still a concern \cite{jamia2020explainability}, as the majority of researchers are more concerned with the performance rather than explainability \cite{uysal2022explainability}. Therefore, a limited number of studies report explainable models, especially in medicine \cite{alsaleh2023explainability}. Explaibanility is necessary because doctors are not likely to use assistive tools and technologies for their decision support if they do not know the rationale behind the inference of an AI assistant \cite{liu2022explainability}. Consequently, the deployment of AI-based decision support interventions in real-world clinical settings is still limited, as reported in a study of the Web Of Science databases from 1995-2019 \cite{guo2020deploymentgap}. 

\subsection{Objectives}

As mentioned above, lab investigations are typically conducted when symptoms of chronic diseases are evident and physicians prescribe them to confirm a diagnosis. Consequently, it gets too late for preventive measures; thus, the cost of healthcare increases exponentially. 

In this study, we utilize routine EHR data from multiple U.S. practices integrated with CureMD’s EMR/EHR system to develop a clinically useful, practical, and explainable predictor for multiple chronic diseases, aiming to predict risks one year in advance. The major contributions of our work include: (1) a comprehensive disease risk surveillance system for eight individual chronic diseases; (2) predictive models that enhance clinical utility by forecasting future risk without the need for lab tests or other data not routinely recorded during clinical encounters; (3) an explainable system that employs novel rule engineering to foster doctor trust and ease integration, thereby bridging the gap between research and practical deployment.

This system empowers physicians to adopt preventive measures like lifestyle changes and dietary recommendations, potentially averting or decelerating chronic diseases. It can enable healthcare systems to allocate resources more effectively, manage the root causes of numerous conditions, reduce treatment costs, and enhance value-based care. Designed to be lightweight, our models integrate seamlessly with existing EMR systems without the need for expensive GPUs. The current study includes the following eight chronic diseases (along with their respective prevalence from CDC's website): Hypertension (48.1\%), Diabetes (16.2\%), Kidney (13.9\%), Heart (4.9\%), COPD (4.6\%), AFIB (1\% - 2\%) \cite{afib_prevalance}, Osteoarthritis (9.75\%), and Hyperlipidemia (11.5\%).


\section{Data Overview}
Anonymized and de-identified EHR/EMR data of patients, in compliance with HIPAA, were provided by the provider practices of CureMD for this research. Since this is a retrospective study, we could either consider a single time window for each patient which would have limited the number of patients; or a dynamic window for each patient \cite{pungitore2023timewindow},  picking the appropriate encounter given an X-month window to make our feature vector, where X can be 3, 6 or 12 months.

For all diseases, adults greater than or equal to 18 years of age were considered for creating training datasets. The patients' count for each disease before splitting are shown in table \ref{tab : data_overview}

\begin{table}[ht]
    \vspace{-0.5cm}
	\caption{Patient counts for each chronic disease that were eligible for model training, and their respective criteria}
	\label{tab : data_overview}
	\centering
	\begin{tabular}{c|c| c }
		\hline
		Disease & ICD10 codes &  Patient Count \\
		\hline
		Hypertension (HTN)                           & I10-I13 & 54654 \\
		Type 2 Diabetes Mellitus (DM)                & E11     & 31865 \\
		Chronic Kidney Disease (CKD)                 & N18, I12, I13, [E08-E13].22 & 15316\\
		Chronic Ischemic Heart Disease (CHD)         & I25     & 13804 \\
		Chronic Obstructive Pulmonary Disease (COPD) & J41-J44 & 15131 \\
        Atrial fibrillation and flutter (AFIB)       & I48     & 9411  \\
        Osteoarthritis (OA)                          & M15-M19, M20.2, M47         & 33997 \\
        Hyperlipidemia (HLD)                         & E78     & 59370 \\
		\hline
	\end{tabular}
    \vspace{-0.5cm}
\end{table}

\begin{figure}[ht]
    \centering
    \includegraphics[width=\textwidth]{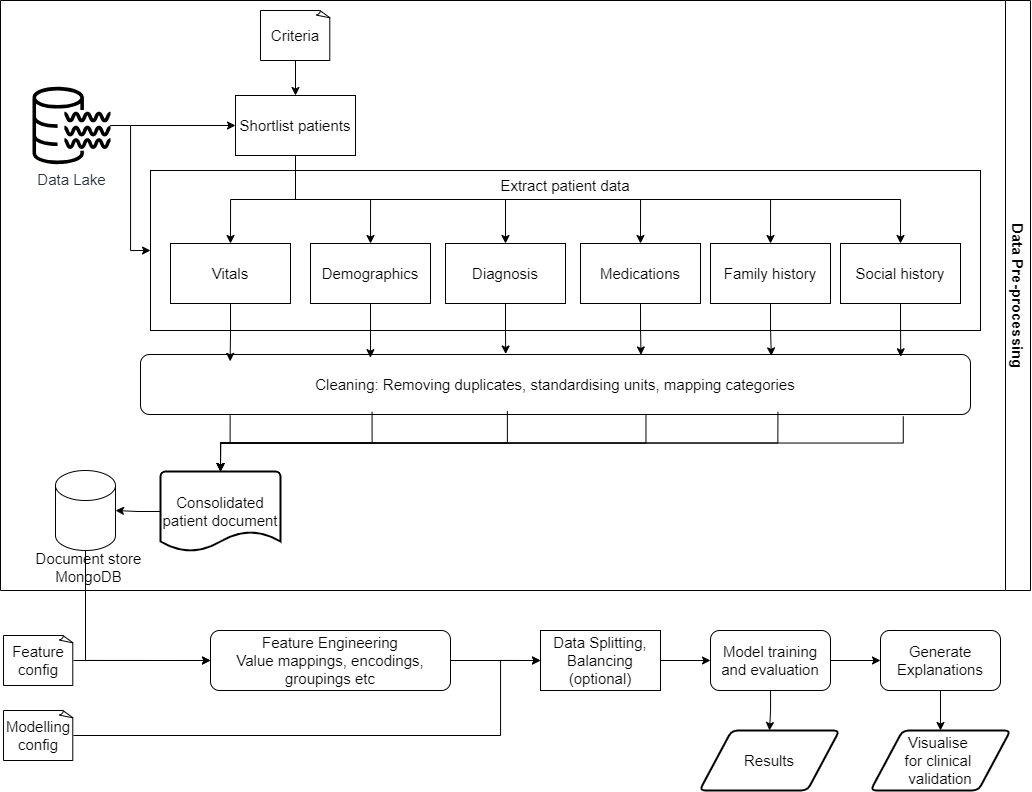}
    \caption{Data preprocessing and model development workflow. Automated pipelines are constructed that require human input via the three configuration steps.}
    \label{fig:preprocessing}
\end{figure}

Extensive pre-processing was done owing to the noise in the raw EMR data. The pre-processing and model development workflows are summarized in Figure \ref{fig:preprocessing}. Clinical validation and evaluations were carried out on the validation set, after which only the feature and modeling configurations would be updated. Three different datasets were prepared for each chronic disease corresponding to the timing windows of 3, 6, and 12 months respectively. Each training dataset was prepared by the following method: the diagnosed patients were sampled by making the feature vector at the earliest encounter within the window; while for normal patients, the feature vector was made at the closest encounter outside the window. 

\section{Methodology}

The study size of each disease was limited by the available data in CureMD’s EHR system. Subsequently, the pre-processing of data also removed some patients. The final sample sizes for each of the eight diseases before splitting are shown in table \ref{tab : data_overview}.

The choice of risk predictors (features) started with a literature review which included identification of risk factors, symptoms, and features used in the previous models. After this, all predictors were used to train the risk surveillance models. We then conducted an investigative study with the panel of clinicians to identify the features that are considered ``dependent or derivative'' of some ``independent'' features. The clinicians were of the opinion that the ``independent features'' provide a more plausible explanation; therefore, ``dependent features'' should be removed from the feature set. Subsequently, we only used the feature sets that were validated by the panel of physicians.
 
The risk surveillance models use features like vitals (such as BMI, MAP etc), demographics (such as age, and race), previous diagnosis, Elixhauser comorbidity groups, social history (such as smoking history), family history, and medications. All features were encoded into nominal or ordinal categorical features - most were simply boolean to mark the presence or absence of diagnoses, medications, and conditions in family history. Continuous features such as age, BMI, and MAP were categorized into ordinal variables, and gender and race were encoded (0 or 1) based on the risk found in the literature (e.g. male was assigned 1 for the Hypertension dataset) while every other feature was binarized (absent/present). Thus, our final feature set contains some ordinal and mostly binary features. Vitals, age, and social history were recorded on the feature vector date. For all other features, any prior feature that was true was also considered true at the time of making the feature vector. 

Missing values were handled in two different ways for two different types of variables. If the value for a vital sign or demographic information e.g. gender was missing, that patient was dropped; while a missing value in diagnosis codes was considered to be marked “absent” in the patient.

\subsection{Analytical Methods} Since we have had a good number of patients for all selected chronic diseases, therefore, the dataset was split into training, validation, and test sets in a 70:10:20 split. The validation set was used for model selection, feature selection, hyper-parameter tuning, and model evaluation.

We trained different types of models (with various combinations of hyperparameters) such as Random Forest, Ada Boosted Models, Light Boosted Machine (LGBM), and Extreme Gradient Boosted Machine (XGBM). Finally, the best classifier was chosen based on the performance metrics on the validation set. The performance was evaluated primarily based on the F1 Score, which is a harmonic mean of Precision and Recall as recommended for the binary classifiers \cite{devries2021f1score}.

Since not-diagnosed/normal patients are present in relatively larger proportions than the diagnosed patients in EHR systems, therefore, under-sampling of normal patients was done, as it increases the prediction accuracy compared to the oversampling that increases the variance \cite{ambesange2020underbalancing}. Under-sampling was done such that a 1:1 ratio between diagnosed and normal patients was ensured (before and after splitting). Apart from the sample size, training and evaluation (validation + test) datasets were created using the same method. 

Each risk surveillance model outputs the probability for diagnosis of an individual chronic disease at a given time point i.e. in summary 24 total predictions are made by our system. We chose this outcome because it represents the risk of developing the disease at that time point. Each model outputs the probability whether a patient might develop the risk of an individual chronic disease. 

\section{Results}\label{sec: results}
The risk surveillance models are packaged into an AI system that was integrated with CureMD’s application. The system helps clinicians to get real-time risk evaluations of chronic diseases for a patient. The best models were selected based on the results obtained on the validation datasets. The performance metrics for 3, 6 and 12 months are tabulated in tables \ref{tab: 3 month result}, \ref{tab: 6 month result}, and \ref {tab: 12 month result} respectively.

    \begin{table}[ht]

    \centering
    \caption{Performance of 3 months models on the test set.}\label{tab: 3 month result}
     \resizebox{\columnwidth}{!}{%
        \begin{tabular}{l |c| c| c| c| c| c| c| c| c}
        \hline
        Disease &   Accuracy & Precision & Recall & NPV & Specificity & AUROC & AUPRC & F1 score \\
        \hline
        Hypertension   & 0.758 & 0.757 & 0.768 & 0.760 & 0.748 & 0.835 & 0.829 & 0.762 \\
        Diabetes       & 0.793 & 0.818 & 0.748 & 0.773 & 0.837 & 0.867 & 0.891 & 0.781 \\
        CKD            & 0.773 & 0.748 & 0.816 & 0.803 & 0.731 & 0.844 & 0.808 & 0.780 \\
        Heart          & 0.784 & 0.763 & 0.808 & 0.806 & 0.761 & 0.861 & 0.825 & 0.785 \\
        COPD           & 0.784 & 0.783 & 0.779 & 0.785 & 0.789 & 0.869 & 0.863 & 0.781 \\
        AFIB           & 0.801 & 0.784 & 0.827 & 0.820 & 0.775 & 0.874 & 0.843 & 0.770 \\
        Osteoarthritis & 0.774 & 0.775 & 0.764 & 0.772 & 0.782 & 0.853 & 0.843 & 0.770 \\
        Hyperlipidemia & 0.761 & 0.766 & 0.757 & 0.757 & 0.766 & 0.835 & 0.817 & 0.761 \\
        \hline
        \end{tabular}
        }
    \end{table}
    \begin{table}[ht]

    \centering
\vspace{-0.5cm}
    \caption{Performance of 6 months models on the test set.}\label{tab: 6 month result}
     \resizebox{\columnwidth}{!}{%
        \begin{tabular}{l |c| c| c| c| c| c| c| c| c}
        \hline
        Disease &   Accuracy & Precision & Recall & NPV & Specificity & AUROC & AUPRC & F1 score \\
        \hline
        Hypertension   & 0.752 & 0.740 & 0.772 & 0.764 & 0.732 & 0.828 & 0.821 & 0.756 \\
        Diabetes       & 0.792 & 0.831 & 0.735 & 0.760 & 0.849 & 0.868 & 0.894 & 0.780 \\
        CKD            & 0.771 & 0.755 & 0.800 & 0.788 & 0.741 & 0.841 & 0.809 & 0.777 \\
        Heart          & 0.770 & 0.742 & 0.817 & 0.803 & 0.725 & 0.842 & 0.802 & 0.778 \\
        COPD           & 0.769 & 0.757 & 0.779 & 0.782 & 0.759 & 0.853 & 0.849 & 0.768 \\
        AFIB           & 0.787 & 0.776 & 0.817 & 0.799 & 0.756 & 0.858 & 0.837 & 0.796 \\
        Osteoarthritis & 0.751 & 0.754 & 0.749 & 0.747 & 0.752 & 0.830 & 0.822 & 0.752 \\
        Hyperlipidemia & 0.754 & 0.754 & 0.757 & 0.754 & 0.752 & 0.820 & 0.796 & 0.755 \\
        \hline
        \end{tabular}
        }
    \end{table}

    \begin{table}[ht]

    \centering
    \caption{Performance of 12 months models on the test set.}\label{tab: 12 month result}
     \resizebox{\columnwidth}{!}{%
        \begin{tabular}{l |c| c| c| c| c| c| c| c| c}
        \hline
        Disease &   Accuracy & Precision & Recall & NPV & Specificity & AUROC & AUPRC & F1 score \\
        \hline
        Hypertension   & 0.738 & 0.728 & 0.754 & 0.748 & 0.721 & 0.813 & 0.805 & 0.741 \\
        Diabetes       & 0.785 & 0.830 & 0.721 & 0.751 & 0.850 & 0.862 & 0.890 & 0.772 \\
        CKD            & 0.764 & 0.747 & 0.796 & 0.783 & 0.733 & 0.835 & 0.805 & 0.771 \\
        Heart          & 0.759 & 0.730 & 0.812 & 0.794 & 0.707 & 0.827 & 0.788 & 0.769 \\
        COPD           & 0.753 & 0.740 & 0.765 & 0.765 & 0.741 & 0.841 & 0.837 & 0.752 \\
        AFIB           & 0.763 & 0.728 & 0.833 & 0.809 & 0.696 & 0.840 & 0.805 & 0.777 \\
        Osteoarthritis & 0.748 & 0.756 & 0.741 & 0.741 & 0.755 & 0.824 & 0.811 & 0.749 \\
        Hyperlipidemia & 0.742 & 0.734 & 0.749 & 0.750 & 0.735 & 0.809 & 0.781 & 0.742 \\
        \hline
        \end{tabular}
        }
    \end{table}

\section{Discussion}

If we take the average of each metric for all eight diseases, across all metrics, the 3 months model shows a superior performance followed by 6 months and 12 months models. This is understandable, as once a patient moves in their disease journey closer to a disease diagnosis, the greater the chances that their feature vector will capture patterns that indicate the onset of the disease. However, the intrinsic value of the 12-month model is the largest, as this helps physicians to consider preventive treatment options, leading to higher value-based care for patients and the healthcare system.

Except for the 12-month hypertension and hyperlipidemia models,  the F1 score for all other chronic diseases is greater than 75\% across 3 models (3, 6, and 12 months), which not only highlights the utility of these risk models, but also underlines the difficulty of predicting hypertension and hyperlipidemia 12 months in advance. Note that the AUROC is greater than 80\% across all chronic diseases and for all 3 time points which makes these models valuable tools in decision support assistance, as confirmed by the panel of physicians. 

\section{Deploying the models}
\begin{figure}
    \centering
    \includegraphics[width=\textwidth]{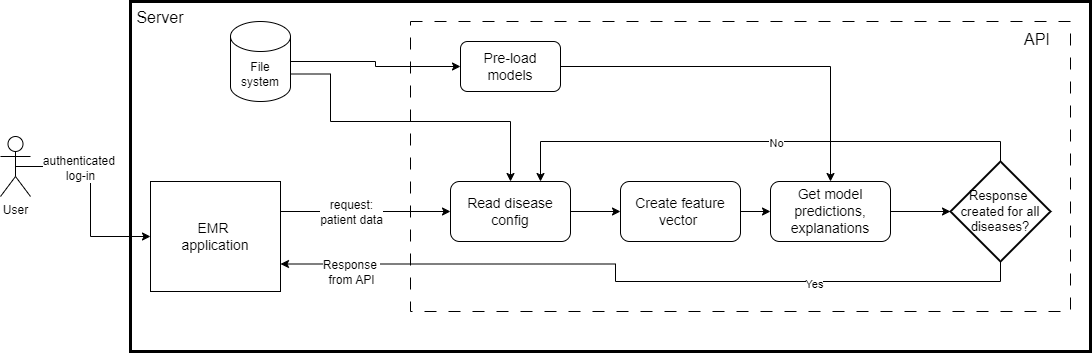}
    \caption{Overall workflow of the prediction API for multiple chronic diseases.}
    \label{fig:api}
\end{figure}

\textbf{Integration with EMR system}.
Figure \ref{fig:api} illustrates the architecture of our current deployment -- an API that is hosted locally with the EMR system to enhance security and data privacy. The secure API retrieves patient's information including diagnoses, vitals, and medications from the EMR system. This data is processed to replicate the steps that are used in the model development, creating disease-specific feature vectors according to the configuration.

One significant advantage of this API framework is its ease of integration across various EMR systems and services. The local hosting ensures that patient information remains within an EMR system's secure enclave, safeguarding patient data's security and privacy. Moreover, the adoption of common information exchange protocols, such as FHIR, facilitates seamless integration with any EMR product or service. Last but not least, the risk surveillance models can run on regular CPUs without the need for expensive computing requirements of GPUs.
\vspace{-0.5 cm}
\vspace{-0.75cm}

\textbf{Explainability}
The interpretability of risk surveillance models, particularly in the healthcare domain, is an important requirement. Clinicians' trust in these models depends on how much they trust the reasons for their inferences. Random forests are typically regarded as black-box models; however, we use the tree structure and model-agnostic methods to better explain inferences made for individual patients.

\begin{figure}[ht]
    \centering
    \includegraphics[width=\textwidth]{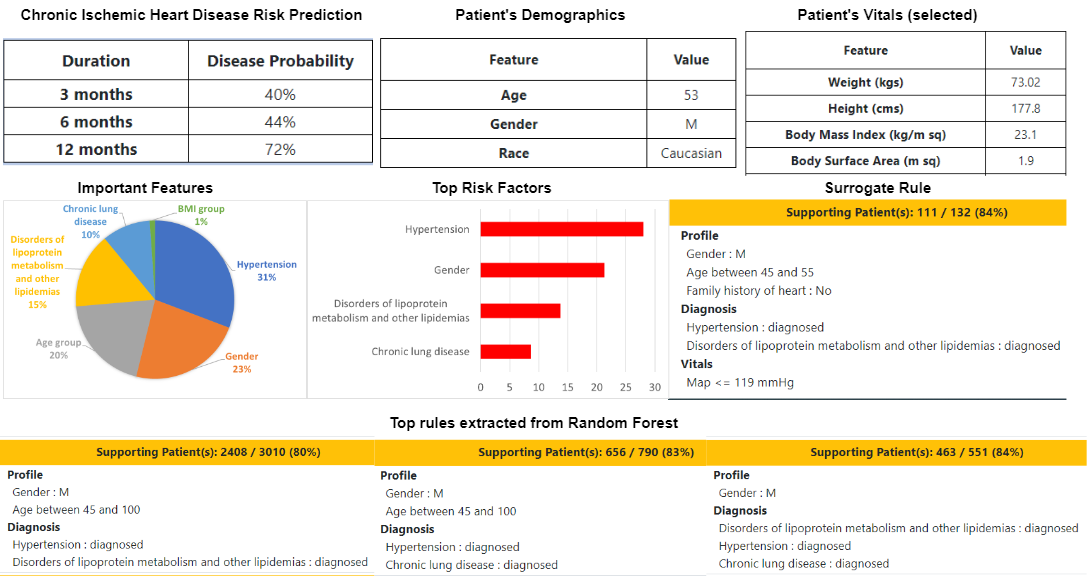}
    \caption{Explanations for CHD prediction for a patient as presented to the user. For this patient, their age and BMI are reducing the risk, while their diagnoses of hypertension (I10), Dyslipidemia (E78), and COPD (J44) are increasing the risk.}
    \label{fig:exlpanations}
\end{figure}

Figure \ref{fig:exlpanations} demonstrates the four types of explanations shown to a physician from within an EMR system that has integrated our risk surveillance models. First, using the Shap library, we determine the important features and present them in a pie chart, showing their relevance in a decision. Second, the risk factors - a subset of the important features that specifically increase the risk of a chronic disease (have a positive contribution) are highlighted in a bar graph. Consequently, a doctor can devise a patient-centered mitigation plan to reduce the risk of each factor, if possible.

Third, we offer two types of rules for a detailed explanation of how our models arrive at their decisions. Initially, we use a surrogate decision tree, an interpretable model trained to approximate the behavior of the random forest. Our surrogate decision trees are limited to a maximum depth of 10, and trained on a subset of the most important features (those contributing 5\% or more globally), ensuring simpler, more accessible explanations. At runtime, we identify the branch of the tree that corresponds to the patient’s feature vector and present its path sequentially. Finally, we try to extract exact rules directly from the random forest.

\vspace{-0.75cm}
\vspace{-0.75cm}
It is important to mention that a forest comprises multiple trees, therefore, we can traverse the path relevant to the current patient in each sub-tree and generate rules in a similar fashion as we would do for a single decision tree. However, this method can yield up to 25 rules for each inference, each containing up to 20 predicates, resulting in a complex set of rules that can be overwhelming and time-consuming for any physician to decipher to make sense of them. Not all splits made during training are critical for explaining decisions; many are one-sided, potentially leading to misinterpretations due to inherent data properties or biases that are influenced by other splits that are at higher levels in the tree. 

To address these issues, we have developed a novel methodology for extracting understandable rules from random forests, illustrated in Algorithm \ref{alg}. Iterating through each tree in a random forest, we identify decision paths leading to the classification of a test instance. For each path (rule), only significant features are kept based on their SHAP values. Rules are then refined based on the bounds of these significant features within the test instance’s neighborhood. These rules are then applied to the entire training set to create a new neighborhood and quantify each rule's support and prediction accuracy.

\begin{algorithm}[ht]
\caption{Extract Simplified Explanation Rules from Random Forest}\label{alg}
\begin{algorithmic}[1]
\Procedure{ExtractRules}{$testInstance, shapImportances, trainingData, randomForest$}
    \State Initialize $explanationRules \gets []$
    \For{each $tree$ in $randomForest$}
        \State $leafID \gets \text{DetermineLeaf}(tree, testInstance)$
        \State $neighbourhood \gets \text{FindNeighbourhood}(trainingData, tree, leafID)$
        \State $branchRule \gets \text{ExtractBranchRules}(tree, leafID)$
        \State $simplifiedRule \gets []$
        \For{each $predicate$ in $branchRules$}
            \State $feature \gets \text{GetFeature}(predicate)$
            \If{$shapImportances[feature] > 0.01$}
                \State $minValue, maxValue \gets \text{GetFeatureRange}(neighbourhood, feature)$
                \State $simplifiedRule.append(feature + \text{"between "} + minValue + \text{" and "} + maxValue)$
            \EndIf
        \EndFor
        \State $newNeighbourhood \gets \text{ApplyRule}(trainingData, simplifiedRule)$
        \State $support \gets \text{CountSamples}(newNeighbourhood)$
        \State $probability \gets \text{CalculateProbability}(newNeighbourhood)$
        \If{$probability \geq 0.5$ and \text{not IsDuplicate}(simplifiedRule, explanationRules)}
            \State $explanationRules.append(\{simplifiedRule, probability, support\})$
        \EndIf
    \EndFor
    \State \textbf{return} $explanationRules$
\EndProcedure

\end{algorithmic}
\end{algorithm}

\section{Conclusion}
Our study introduces a clinically useful risk surveillance system capable of predicting multiple chronic diseases 3, 6, and 12 months in advance. Currently, the models have been evaluated using datasets exclusively from CureMD partner practices. For broader applicability, further evaluation with diverse electronic health record (EHR) datasets is essential to verify the models' generalizability.

Moving forward, we aim to broaden the scope of diseases covered and extend the prediction timeline, enhancing the system's utility for real-world applications. We also plan to enlarge our panel of clinicians to strengthen the validation process, ensuring more robust and comprehensive model assessments.

In summary, we have developed a practical, explainable, and clinically useful disease risk surveillance system for the early prediction of eight chronic diseases. This system addresses a significant gap in healthcare by enabling clinicians to implement preventive measures, thus offering quality-based care that potentially reduces the need for treatments with significant side effects. An emphasis on lifestyle and dietary modifications, inspired by the predictions and explanations from our system would mitigate healthcare costs and allow savings to be reallocated. Additionally, large healthcare entities like the NHS could utilize the identified risk factors to generate Real-World Evidence (RWE) of chronic disease onset, aiding in strategic healthcare planning and budget allocation. These eight chronic diseases represent more than 80\% of the patient population suffering from chronic conditions, underscoring the substantial impact of our study.

\end{document}